\documentclass[letterpaper]{article} 
\usepackage{aaai23-r2hcai}  
\usepackage{times}  
\usepackage{helvet}  
\usepackage{courier}  
\usepackage[hyphens]{url}  
\usepackage{graphicx} 
\usepackage{natbib}  
\usepackage{caption} 
\usepackage{algorithm}
\usepackage{algorithmic}

\usepackage{amsfonts}

\usepackage{newfloat}
\usepackage{listings}
\DeclareCaptionStyle{ruled}{labelfont=normalfont,labelsep=colon,strut=off} 
\floatstyle{ruled}
\newfloat{listing}{tb}{lst}{}
\floatname{listing}{Listing}
\usepackage{fancyhdr}
\fancypagestyle{firstpagehf}
{
   \fancyhf{}
   \fancyhead[HC]{The CONCATENATE workshop at the ACM/IEEE International Conference on Human-Robot Interaction 2023}
   \fancyfoot[C]{\thepage}
}

\title{How much informative is your XAI? A decision-making assessment task to objectively measure the goodness of explanations}
\author{
    Marco Matarese$^{1,2}$ \quad
    Francesco Rea$^2$ \quad
    Alessandra Sciutti$^2$
}
\affiliations {
    \textsuperscript{\rm 1} University of Genoa \quad
    \textsuperscript{\rm 2} Italian Institute of Technology\\
    {marco.matarese, francesco.rea, alessandra.sciutti}@iit.it
}

\usepackage{bibentry}

\begin{document}
\thispagestyle{firstpagehf}
\maketitle

\begin{abstract}
There is an increasing consensus about the effectiveness of user-centred approaches in the explainable artificial intelligence (XAI) field.
Indeed, the number and complexity of personalised and user-centred approaches to XAI have rapidly grown in recent years.
Often, these works have a two-fold objective: (1) proposing novel XAI techniques able to consider the users and (2) assessing the \textit{goodness} of such techniques with respect to others.
From these new works, it emerged that user-centred approaches to XAI positively affect the interaction between users and systems.
However, so far, the goodness of XAI systems has been measured through indirect measures, such as performance.
In this paper, we propose an assessment task to objectively and quantitatively measure the goodness of XAI systems in terms of their \textit{information power}, which we intended as the amount of information the system provides to the users during the interaction.
Moreover, we plan to use our task to objectively compare two XAI techniques in a human-robot decision-making task to understand deeper whether user-centred approaches are more informative than classical ones.
\end{abstract}

\section{Introduction}
The number and complexity of user-centred approaches to XAI have been increasing in recent years \cite{williams2021towards}. 
The application contexts of such approaches are various. They go from computer applications aiming at providing personalised teaching \cite{embarak2022internet}\cite{cohausz2022towards} to human-robot interaction (HRI) contexts in which the robot maintains users' models to provide explanations tailored to them \cite{matarese2021user}\cite{stange2022self}.

These works show a clear trend: user-centred XAI positively affects the interaction between users and systems. It brings to higher users' willingness to reuse the system (\textit{e.g.}, with recommendation systems \cite{conati2021toward}), robots' persuasiveness during human-robot decision-making tasks and human-AI teams performance \cite{schemmer2022meta}.

One of the reasons why the XAI research field is moving toward user-centred approaches is to improve the \textit{goodness} of the explanations.
Alongside producing personalised XAI, recent works also aim to evaluate the goodness of the explanation produced.
What is meant for ``XAI goodness'' is still under debate; nonetheless, there is a broad consensus about the dependency of this concept on the application context. 
Unfortunately, most of these evaluations rely on subjective or indirect measurements (\textit{e.g.}, questionnaires or performance).

However, the main aim of an XAI system is to provide the user with information about the functioning of the underlying AI model.
So far, the amount of information an XAI system provides has been assessed through indirect measures, such as users’ ability to simulate the AI behaviour. 
Indeed, recent surveys \cite{mohseni2018survey} and systematic reviews  \cite{nauta2022anecdotal} highlight the need for more objective and quantitative measures to assess the goodness of XAI techniques.

In our opinion, it is worth taking a step back and measuring the goodness of XAI systems from the information they can provide to users. 
Although measuring how much information a system can generate could be challenging, we can focus on the amount of \textit{new knowledge} that such flow of information creates in the users’ mental models. 
Hence, if we let only non-expert users interact with XAI systems, we can state that the knowledge they acquired arose from the interaction with the system.
Finally, if such knowledge is quantifiable, we can assess how much the system has been informative.

In this paper, we aim to propose an assessment task to objectively and quantitatively measure the goodness of XAI systems during human-AI decision-making tasks intended as how much they are informative to non-expert users.
Moreover, we plan to use such a task to study the effectiveness of user-centred XAI in HRI scenarios.
In particular, we are interested in understanding whether a user-centred XAI approach is more informative than a classical one.

\section{Related work}
Several works assessed XAI systems’ properties through user studies. 
The more interesting ones, in our opinion, are addressed to non-expert users \cite{janssen2022will}. 
These latter used tasks that only some people know about. 
For example, authors in \cite{lage2019evaluation} used two application domains: an artificial alien’s food preferences based on the meals' recipes, and clinical diagnosis. 
Those in \cite{wang2021explanations} also used two peculiar decision-making tasks: a recidivism prediction and a forest cover prediction task. 
Further, authors in \cite{van2021evaluating} used a diabetes self-management use-case where naive users and the system had to find the optimal insulin doses for meals. 
Finally, both \cite{goyal2019counterfactual} and \cite{wang2020scout} used image classification tasks in which participants were asked to recognise two bird species.

Several works regarding the assessment or comparison of XAI methods tend to define their own measure of goodness \cite{van2021evaluating}\cite{lage2019evaluation}.
However, a method has been proposed to objectively measure the \textit{degree of explainability} of the information provided by an XAI system \cite{sovrano2022quantify}. 
Moreover, the authors in \cite{holzinger2020measuring} proposed the System Causability Scale to measure the quality of the explanations based on their notion of causability \cite{holzinger2019causability}.
Finally, the authors in \cite{wang2022effects} proposed a different point of view regarding assessing XAI systems' goodness by comparing several types of XAI in different application contexts with respect to three desiderata: to improve people’s understanding of the AI model, help people recognise the model uncertainty, and support people’s calibrated trust in the model.

Recent works regarding user-centredness focus on users’ trust towards the system and highlight context-awareness and personalisation as main approaches to user-centredness \cite{williams2021towards}. 
Personalisation in XAI has also been implemented by exploiting the users’ personality traits and correlating them with users' preferences or behaviours \cite{bockle2021can}\cite{martijn2022knowing}.
Instead, authors in \cite{bertrand2022cognitive} reviewed a relevant corpus of literature to understand which human biases researchers reflect in their XAI methods (also without noticing). 

Most of the works in the XAI field with user studies regard decision-making \cite{wang2021explanations} or classification tasks \cite{goyal2019counterfactual}.
The rationale behind this choice is the promise of better performance when coupling human users with expert AI systems \cite{wang2022effects}. 
Such a promise is almost always kept, although a few studies show that team performance decreases when using some form of XAI \cite{schemmer2022meta}.
Indeed, some authors highlighted that AI advice is not always beneficial mainly because humans have shown to be unable to ignore incorrect AI advice \cite{schemmer2022should}\cite{ferreira2021human}\cite{janssen2022will}.


\section{Motivation and hypotheses}
So far, researchers have measured the goodness of an XAI system used in human-AI decision-making tasks by relying on indirect measures, such as team performance, systems' persuasiveness, or users' ability to predict the AI's decisions.
This means that no objective and quantitative measures have been used to state how good XAI systems are \textit{per se}.
This lack contributes to the impossibility of rigorously comparing two XAI strategies regardless of their particular application context.
Since all the XAI systems have been examined with respect to their application scenario, it is also hard to generalise the results of such research.

We need to introduce an objective and quantitative assessment task to measure one of the most-ignored-so-far factors of XAI: its \textit{information power}. 
With the term information power, we mean the amount of information that an XAI system provides about: (1) underlying AI models' (general) functioning, (2) reasons behind a particular model's choice, or (3) what the system would do in other circumstances.

Under the assumption that the goodness of an XAI system reflects the accuracy of the users' mental models about the underlying AI system \cite{hoffman2018metrics}, we want to understand deeper why user-centred XAI is more effective than those approaches that do not take the users in consideration. Our experimental hypothesis regards the users' level of involvement in the explanation generation process.
In particular, the more users' are involved in the explanation generation process:
\begin{itemize}
     \item[H1] the more the XAI system has more information power (resulting in more accurate users' mental models of the AI).
    \item[H2] the more they become independent of the robot's suggestions and explanations.
    \item[H3] the higher the users' willingness to reuse the XAI system.
    \item[H4] the more positive the users' feelings toward the robot.
\end{itemize}

\begin{figure}[t]
    \centering
    \includegraphics[width=0.5\textwidth]{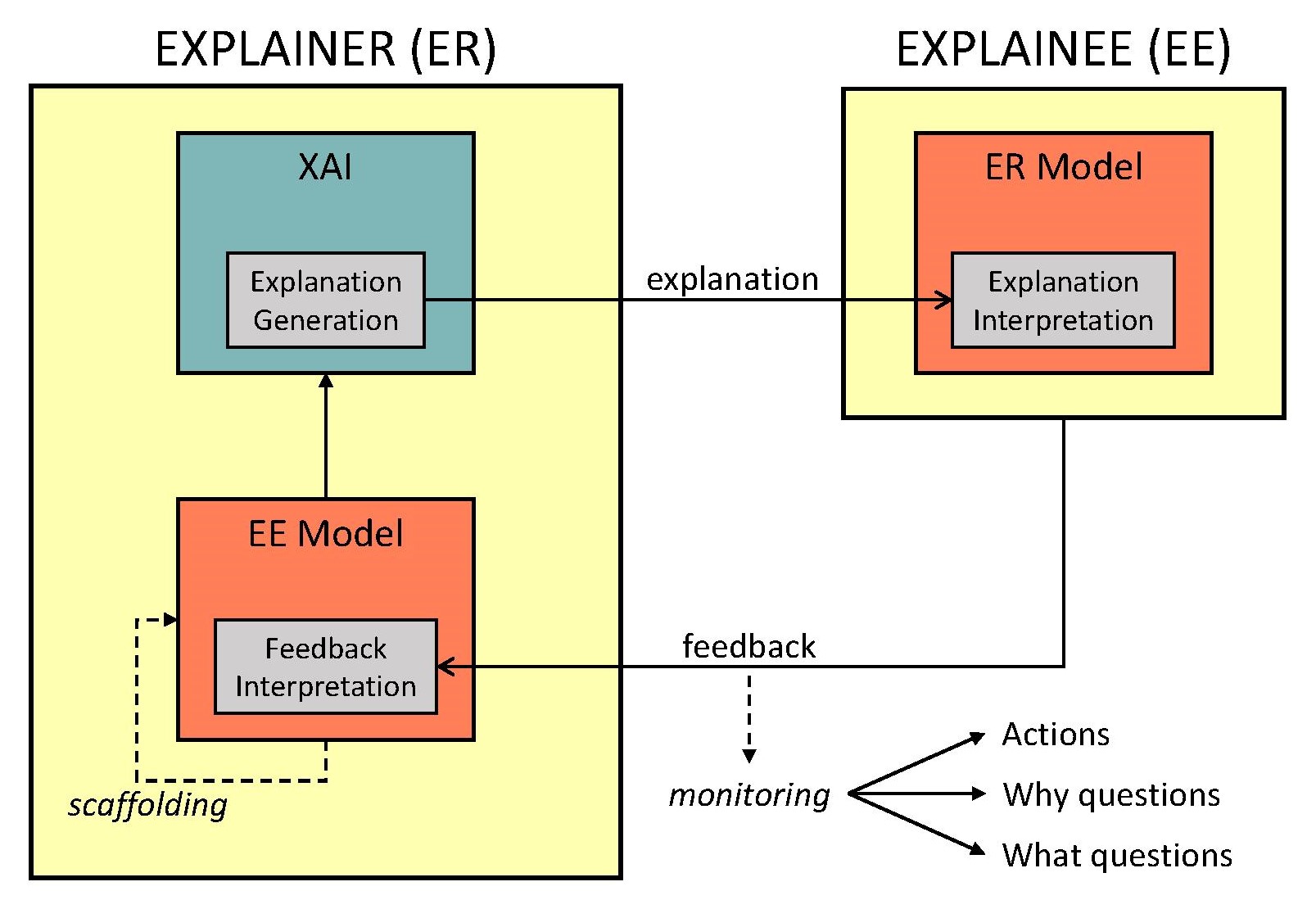}
    \caption{The co-constructing approach to the explaining process, as in \cite{rohlfing2020explanation}. The actors' explanation behaviours adapt by monitoring and scaffolding each other. In our setting, the ER monitors the EE through feedback about their actions and questions.}
    \label{fig:cocostruction}
\end{figure}

\begin{figure*}[t]
    \centering
    \includegraphics[width=.8\textwidth]{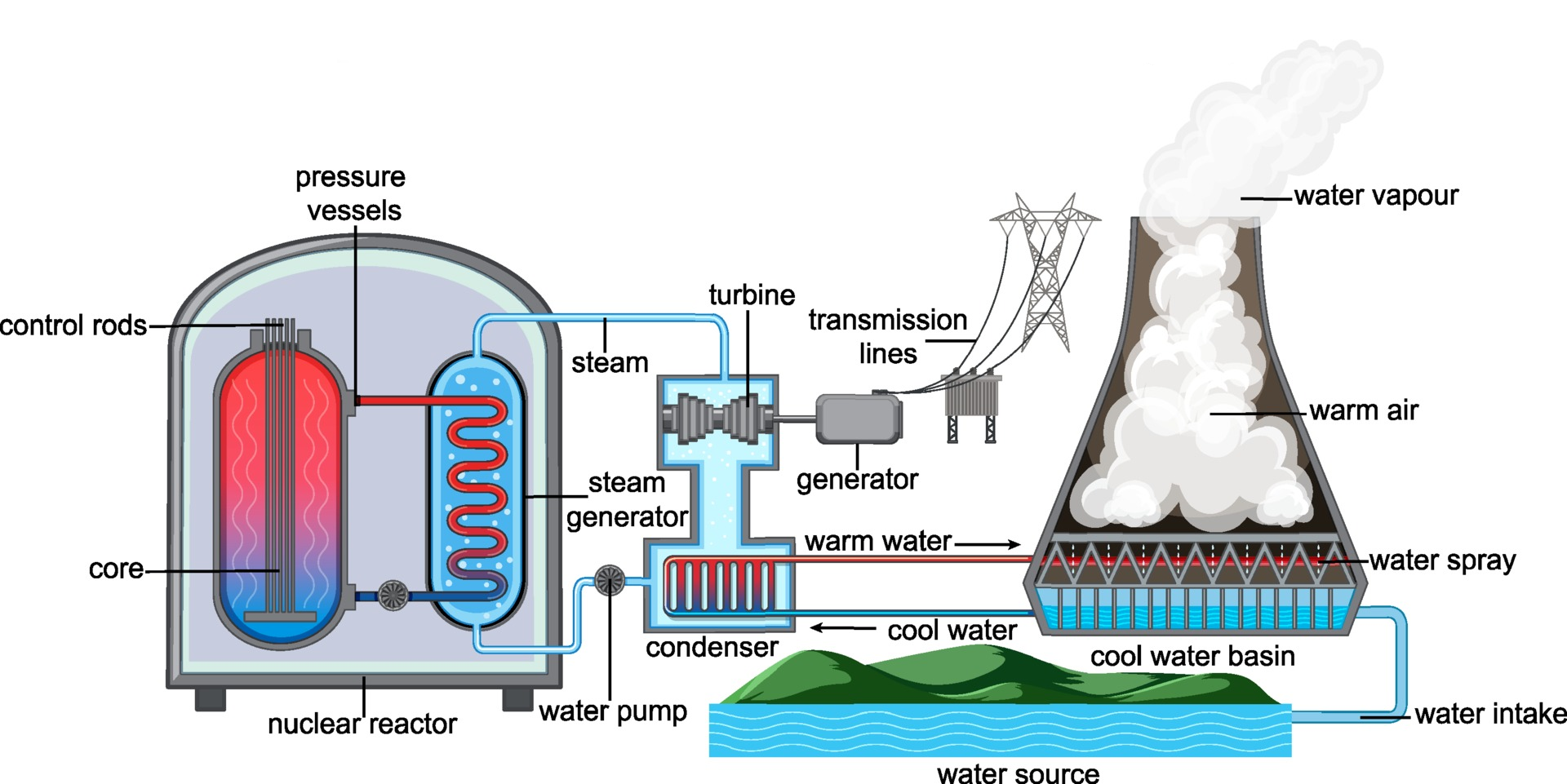}
    \caption{The schema of a pressurised water reactor that we implemented in our simulated environment. This schema considers only the control rods, but we also allow the users to manage the other three types of rods: the fuel, sustain and regulatory ones.}
    \label{fig:pwr}
\end{figure*}

\section{Methods}
We are interested in introducing an objective and quantitative assessment task to measure the goodness of XAI systems without relying only on indirect measures.
Hence, we plan to measure the XAI systems' information power directly and validate the task through a user study.

\subsection{The task}
\label{sec:task}
The assessment consists of a decision-making task where users can interact with a control panel to perform actions in a simulated environment (see Section \ref{sec:env}).
During the task, users can interact with an expert AI (in our case, with the humanoid robot iCub) by asking \textit{what} it would do, and its XAI system by asking \textit{why} it would do that. 
Apart from the instructions about interacting with the control panel and the robot, users start the task knowing nothing about it.

In a fixed amount of time, the users have to perform actions and interact with the robot to discover (1) which is the task at hand (\textit{e.g.}, what is the goal of the task), as well as (2) the rules that the AI model uses to make its actions and, since the robot is an expert agent, (3) the rules that govern the simulated environment.

\subsubsection{Interaction modalities}
The roles between the human-robot team and their interaction modalities are simple.
The robot can not perform actions, but its role is limited to assisting users during decision-making.
However, the robot can not take the initiative in giving suggestions either, but it always replies to users' questions.
Thus, only the users can interact with the control panel and act in the simulated environment.

We expect the following HRI modalities. 
If the user already knows what to do, they act on the control panel. 
Otherwise, they ask the robot what it would do; then, the robot answers this \textit{what} question saying what action it would do in the current scenario.
If the user understands the robot's reasons, they act on the control panel (not necessarily the suggested one).
Otherwise, the user asks the robot why it would do that.
At this point, the robot answers this \textit{why} question by explaining as justification for its suggestion.
All that remains for the user is to act on the control panel.

\subsubsection{Characteristics of the task}
We need non-expert users to consider the information passed through the interaction as new information.
Taking it to extremes, we consider only participants without knowledge about the task and its underlying rules. 
For this reason, we implemented a simulation of a nuclear power plant management task (Figure \ref{fig:pwr}).
We chose this kind of task because it met all the requirements we needed: it is challenging and captivating for non-expert users, simple rules govern it, an AI model can learn those rules and, usually, people know nothing about the functioning of nuclear power plants.

The main objectives of the task (which we hide from users) are to generate as much energy as possible and maintain the system in an equilibrium state.
The features of the environment are subject to rules and constraints, which we can summarise as follow:
\begin{itemize}
    \item[$\cdot$] Each action corresponds to an effect on the environment: thus, a change of its features' value.
    \item[$\cdot$] Several preconditions must be satisfied to start and continue nuclear fission and produce energy.
    \item[$\cdot$] Some conditions irremediably damage the plant.
    \item[$\cdot$] The task is divided into steps of fixed time duration (\textit{e.g.}, 10 seconds), in which the users can interact with either the robot or the control panel.
\end{itemize}

\subsection{The simulated environment}
\label{sec:env}
\subsubsection{Features of the environment}
Our simulated power plant is composed of 4 continuous features: 
\begin{itemize}
	\item[$\cdot$] The pressure in the reactor's core.
	\item[$\cdot$] The temperature of the water in the reactor
	\item[$\cdot$] The amount of water in the steam generator
	\item[$\cdot$] The reactor's power.
\end{itemize}
Furthermore, the power plant has four other discrete features that regard the reactor's rods: security rods, fuel rods, sustain rods, and regulatory rods. 
The firsts two have two levels: up and down. Instead, the latter two have three levels: up, medium, and down.

The reactor power linearly decreases over time for the effect of the de-potentiation of the fuel rods. 
Hence, the reactor's power depends on the values of the environment's features and whether nuclear fission is taking place.
Moreover, the energy produced at each step is computed by dividing the reactor's power by 360, which is the power that a 1000MW reactor without power dispersion produces in 10 seconds.

\subsubsection{Actions to perform on the environment}
The actions that the user can perform (12 in total) go from changing the position of the rods to adding water to the steam generator or skipping to the next step.
All those actions change the value of 3 parameters - $T$, $P$, and $L$ - which correspond to the water's temperature in the core, the core's pressure, and the water level in the steam generator, respectively. 
The setting of the rods determines the entity of the feature updates; such updates are performed at the end of each step, right after the users' action.
For example, if the safety rods are lowered in the reactor's core, the nuclear fission stops; thus, $T$ and $P$ decrease until they reach their initial values, and the water in the steam generator remains still.
On the other hand, if nuclear fission occurs and the user lowers the regulatory rods, the fission accelerates. This acceleration consumes more water in the steam generator, raising the core's temperature and pressure more quickly, but also raising the reactor's power and the electricity produced.
If the users do not act within the time provided for each step, the application automatically chooses a \textit{skip} action, which applies the features' updates based on the setting of the rods at hand.

\subsection{The robot's AI}
\label{sec:ai}
Regarding the robot's AI model, we trained a deterministic decision tree (DT) using the Conservative Q-Improvement learning algorithm \cite{roth2019conservative}, which allowed us to train the DT using a reinforcement learning strategy.
Instead of extracting the DT from a more complex ML model \cite{vasilev2020decision}\cite{xiong2017learning}, we used this learning strategy to simplify the translation from the AI to the XAI without losing performance.
The robot uses this expert DT to choose its action: it can perform each of the 12 actions based on the eight environment's features.

Starting from its root node, the DT is queried on each of its internal nodes - which represent binary splits - to decide which of the two sub-trees continue the descent. 
Each internal node regards a feature $x_i$ and a value for that feature $v_i$: the left sub-tree contains instances with values of $x_i \leq v_i$, while the right sub-tree contains instances with values of $x_i > v_i$ \cite{buhrman2002complexity}.

The DT's leaf nodes represent actions; in the implementation of Roth \textit{et al.} \cite{roth2019conservative}, they are defined with an array containing the actions' expected Q-values: the greater Q-value is associated with the most valuable action.
This way, the DT can be queried by users with both \textit{what} and \textit{why} questions.
To answer a \textit{what} question, we only need to navigate the DT using the current values of the environment's features and present the resulting action to the user.

\begin{figure}[t]
    \centering
    \includegraphics[width=0.4\textwidth]{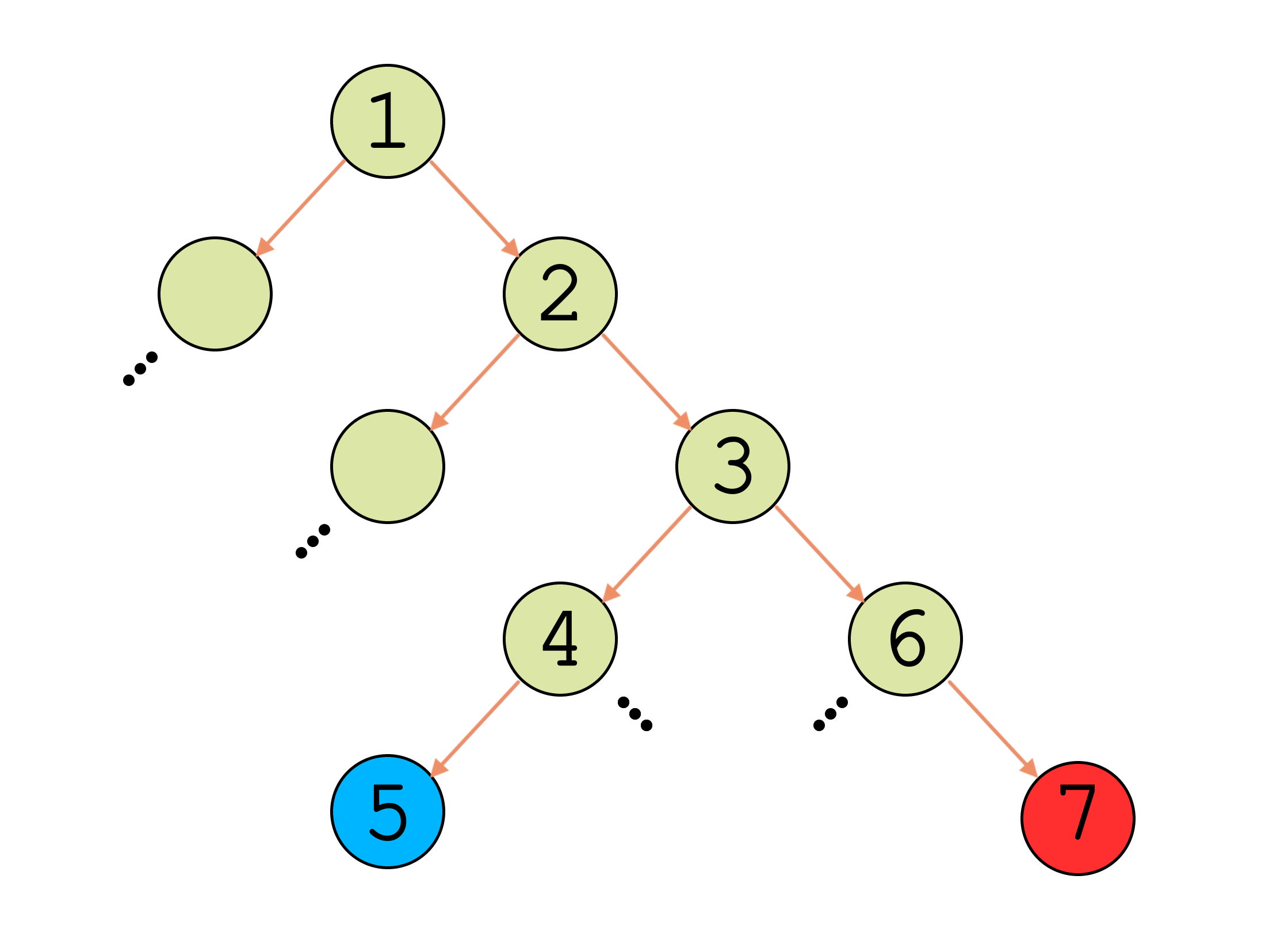}
    \caption{An example of DT where the leaf nodes 5 and 7 are the robot's suggestion and the predicted user's action, respectively. Let us assume that node 1 has already been used: then, the \textit{classical} XAI selects node 2 for the explanations since it is the most unused relevant node. The \textit{user-aware} XAI selects instead node 3 because it represents a perfect counterfactual explanation for fact 5 and foil 7.}
    \label{fig:dt}
\end{figure}

\subsection{The robot's XAI}
As we already saw in Section \ref{sec:task}, answering \textit{why} questions regards the robot XAI.
Since the AI model to explain is already transparent \cite{adadi2018peeking}, it also constitutes the XAI model.
We can use DTs to provide explanations simply by using one (or more) of the feature values we encounter during the descent.

As we have seen in Section \ref{sec:ai}, during the DT descent, we encounter a set of split nodes defined by a feature $x_i$ and a value $v_i$; the direction of the descent tells us if the current scenario has a value of $x_i \leq v_i$ or $x_i > v_i$.
Each of those inequalities can be used to provide an explanation that can help user to relate actions with specific values of the environment's features.
In our case, an explanation for the action "add water to the steam generator" could be "because the water level in the steam generator is $\leq 25$, which is dangerously low.

Which of the features we encounter during descent to use is a problem called ``explanation selection''. 
In our case, the explanation selection depends on the XAI strategy we want to test.
For example, classical approaches use the most relevant features (in terms of the Gini index, information gain, or other well-established measures \cite{stoffel2001selecting}).
We plan to objectively and quantitatively compare two XAI strategies at first glance: a classical approach vs an \textit{user-aware} one.

The \textit{classical} XAI explains using only the AI outcomes and environment's states. 
In particular, it justifies the robot's suggestions by using the most relevant features, which are the first ones in the DT's structure (see \cite{roth2019conservative}).
The system tries to give always different explanations by taking track of the DT's node already used and preferring to use the others in decreasing the level of relevance.

On the other hand, the \textit{user-aware} XAI approach, through monitoring and scaffolding (Figure \ref{fig:cocostruction}), maintains a model of the users to take track of their previous actions and predict their future behaviour \cite{rohlfing2020explanation}. 
Hence, the predicted users' actions can be used to produce counterfactual explanations: the \textit{fact} (the outcome to explain) will be the robot's suggestion; in contrast, the \textit{foil} (the expected outcome) will be the predicted users' action.
Figure \ref{fig:dt} shows an example of these two approaches.

\subsection{Measuring the explanations' goodness}
The assessment of the model's information power involves the interaction between non-expert users and the system itself.
During this assessment, the users aim to learn as many system rules as possible.
Thus, we need to collect measures for each rule and combine them to obtain the model's information power.
The general assessment steps are the following:
\begin{enumerate}
    \item To quantify how many rules regard each feature and define a method for measuring the number of learned rules relative to each feature.
    \item To quantify how much informative weight each feature has. For the sake of simplicity, we can assume that they have the same informative weight (equal to $\frac{1}{k}$, where $k$ is the number of features). The features' informative weights must respect the following rule: $\forall j \in \{1, ..., k\}, \sum_j \gamma_j = 1$. The features' informative weight is intended to describe how difficult it is to understand the rules regarding such features.
    \item To measure the model's informative power for each user, obtaining a measure (and a set of secondary descriptive measures) for each of them.
    \item To average those measures obtaining the final results; the secondary descriptive measures could also have valuable meaning.
\end{enumerate}

Hence, if $k \in \mathbb{N}$ is the number of the model's features, $\gamma_j \in [0, 1]$ is the informative weight of the feature $j$, $n_j^r \in \mathbb{N}$ is the number of rules regarding the feature $j$, $n_j^{lr(i)} \in \mathbb{N}$ is the number of rules regarding the feature $j$ learned by the user $i$, and $a_m$ is the accuracy of the AI model $m$, then the informative power of the model $m$ for the user $i$ is computed as follows:
\begin{equation}
    IP_i(m) = a_m \sum_{j=1}^{k} \gamma_j \left( \frac{n_j^{lr(i)}}{n_j^r} \right) \in [0, 1].
\end{equation}
Thus, if $n^p$ is the number of users who took part in the assessment, the information power of the model $m$ is 
\begin{equation}
    IP(m) = \frac{1}{n^p} \sum_{i=1}^{n^p} IP_i(m) \in [0, 1].
\end{equation}

Apart from the number of rules regarding each feature, the most delicate aspect of the assessment regards the definition of the features' information weights. 
We suggest at least two ways to set them: to make them all equal or to define the weights using experimental data.
The former approach depends on the task at hand. For our task (Section \ref{sec:task}), we set the features' information weight by normalising the number of interactions with the system that users needed to understand those features.
To state about which feature the interaction regarded, we plan to proceed as follows: if the interaction stopped at the request for suggestions, then the interaction regards the feature on which the action suggested effects; otherwise, if the interaction continues with a request for an explanation, then the interaction regards the feature contained into the explanation.

\subsubsection{Experimental measures}
During our task, we plan to collect several quantitative measures to compute the model's information power as defined above:
\begin{itemize}
    \item[(M1)] Measures of performance, such as the users' final score.
    \item[(M2)] Measures of rules understanding, such as the number of task rules learned, the number of requests and interactions users needed to learn such rules, and the number of correct answers to the post-experiment test.
    \item[(M3)] Measures of generalisation, such as the number of correct answers to \textit{what-if} questions about the robot's decisions in particular states of the environment.
\end{itemize}
Moreover, we collect some subjective measures:
\begin{itemize}
    \item[(M4)] Measures of satisfaction, such as users' satisfaction level about the explanations and the interaction.
    \item[(M5)] Measures of robot perception, such as users' feelings towards the robot and perception of it.
\end{itemize}

\section{Discussion}
The assessment task we propose satisfies properties unique to the XAI research field.
Firstly, it focuses on the information power of the XAI system intended as the amount of information it can give to the user.
Then, it defines the goodness of the XAI system as a function of its information power.
Secondly, it allows for an objective and quantitative analysis of such goodness.

The most critical requirement of our assessment task is user interaction. 
It allows for a two-way, possibly iterative, interaction with the users.
In particular, it allows the users to query the system by asking what it would do in a specific situation and why. 
Consequently, the XAI system should be able to answer both \textit{what} and \textit{why} questions to exploit the full potential of the assessment task.

Another essential factor of our task is that we can easily generalise it.
To generalise our task, we need the following:
\begin{itemize}
    \item[$\cdot$] A decision-making task with characteristics similar to the ones listed in Section \ref{sec:task}.
    \item[$\cdot$] An expert AI and several non-expert users.
    \item[$\cdot$] The approaches to compare could be XAI algorithms or HRI dynamics.
    \item[$\cdot$] At least one quantitative measure about users' understanding of the task: if two or more, a method to compact them into a single measure is also needed.
    \item[$\cdot$] At least one quantitative measure about users' ability to generalise to unseen scenarios: if two or more, a method to compact them into a single measure is also needed.
\end{itemize}

An assessment task with those characteristics is flexible enough to test different AI models and XAI techniques as long as they allow interaction between the user and the system.
Moreover, it can be used to objectively compare two or more HRI approaches to test whether HRI dynamics ease the users in understanding the robot's suggestions and explanations.
Indeed, in our future work, we plan to use this task to test whether interacting with a social robot makes users more receptive to learning than interacting with a non-social one.

\bibliography{main.bib}

\end{document}